\documentclass[conference]{IEEEtran}
\IEEEoverridecommandlockouts
\usepackage{cite}
\usepackage{amsmath,amssymb,amsfonts}
\usepackage{algorithmic}
\usepackage{graphicx}
\usepackage{textcomp}
\usepackage{xcolor}
\usepackage{hyperref}
\hypersetup{
	colorlinks=true,
	linkcolor=blue,
    filecolor=magenta, 
    urlcolor=cyan
}
\def\BibTeX{{\rm B\kern-.05em{\sc i\kern-.025em b}\kern-.08em
    T\kern-.1667em\lower.7ex\hbox{E}\kern-.125emX}}
\begin{document}

\title{Discrete Cosine Transform in JPEG Compression}

\author{\IEEEauthorblockN{Jacob John}
\IEEEauthorblockA{\textit{School of Computer Science and Engineering (SCOPE))} \\
\textit{Vellore Institute of Technology}\\
Vellore, India \\
jacob.john2016@vitalum.ac.in}
}

\maketitle

\begin{abstract}
Image Compression has become an absolute necessity in today’s day and age. With the advent of the Internet era, compressing files to share among other users is quintessential. Several efforts have been made to reduce file sizes while still maintain image quality in order to transmit files even on limited bandwidth connections. This paper discusses the need for Discrete Cosine Transform or DCT in the compression of images in Joint Photographic Experts Group or JPEG file format. Via an intensive literature study, this paper first introduces DCT and JPEG Compression. The section preceding it discusses how JPEG compression is implemented by DCT. The last section concludes with further real-world applications of DCT in image processing.
\end{abstract}


\section{Introduction}
Two-dimensional images stored in digital format are a collection of millions of pixels each represented as a combination of bits. These images are referred to as raster or bitmapped images, as opposed to vector images that use a mathematical formula to create geometric objects.  As the demand for better quality images and videos increases, efforts have been made to increase the resolution images are being stored in \cite{b1}\cite{b2}\cite{b3}.

For example, a 4k image, i.e., $3840 \times 2160$ pixels in resolution, stored in RAW/DNG 16 bits/pixel format would require around 23 MB to store \cite{b4}. This would mean 45,000 images would occupy roughly 1 TB of storage space, increasing transmission bandwidth and transmission time when sharing several images at once. Furthermore, hosting images on web servers would also cost storage space while also increasing load times, making customer experiences suboptimal \cite{b5} 

Image compression eradicates this need for large storage space by offering efficient solutions for sharing, viewing and archiving a large number of images. Some generic image formats that offer image compress are JPG, TIF, GIF, and PNG \cite{b6}.

\subsection{Discrete Cosine Transform}

During the past decade, the Discrete Cosine Transforms or DCT, has found its application in speech and image processing in areas such as compression, filtering, and feature extraction. Using DCT, an image can be transformed into its elementary components \cite{b7}. DCT uses a sum of cosine functions oscillating at different frequencies to express a sequence of finitely many discrete and real data points with even symmetry \cite{b8}. This can be expressed as equation \eqref{eq1} that consists of a set of basis vectors that are sampled cosine functions.

\begin{multline}
F[k,l]=\alpha(k)\alpha(l)\sum_{m=0}^{N-1} f(m,n)\cos\left[\frac{(2m+1)\pi k}{2N}\right] \\
\cos\left[\frac{(2n+1)\pi l}{2N}\right]\label{eq1}
\end{multline}
where,
\begin{equation}
\alpha(k)=\begin{cases} 
\sqrt{\frac{1}{N}}\:\text{if}\:k = 0\\
\sqrt{\frac{2}{N}}\:\text{if}\:k \neq 0\\
\end{cases}\label{eq1}
\end{equation}

Furthermore, the set of basis vectors given by $\cos\left[\frac{(2m+1)\pi k}{2N}\right]$ and $\cos\left[\frac{(2n+1)\pi l}{2N}\right]$ are a class of discrete Chebyshev polynomials \cite{b9}. Thus, these basis vectors can be defined recursively and composes a polynomial sequence.

The 2D DCT for a signal $x[n]$ of length $N$, is given by $X[k]$ in equation \eqref{eq2}.

\begin{equation}
X[k]=\sum_{n=0}^{N-1}x[n]e^{-\frac{2\pi kn}{N}}\label{eq2}
\end{equation}

where $k$ varies between 0 to $N-1$.

\begin{figure}[htbp]
\centerline{\includegraphics[width=0.4\textwidth]{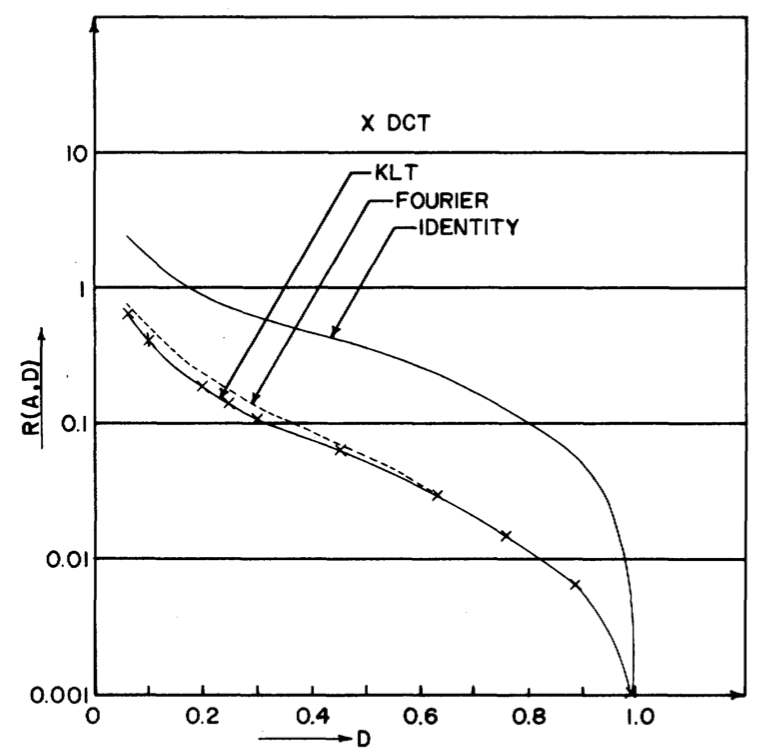}}
\caption{Rate-distortion criteria of various transforms for $M = 16$ and $\rho=0.9$ . \cite{b7}}
\label{fig1}
\end{figure}

Due to its performance with respect to the rate-distortion criterion defined in \cite{b10} and results given in Figure 1, DCT is said to be analogous to the Karhunen-Loève transform (KTL) for first-order Markov stationary random data \cite{b11}. Figure \ref{fig1} also shows how close the curve for DCT is to Karhunen-Loève transform.

\begin{figure}[htbp]
\centerline{\includegraphics[width=0.4\textwidth]{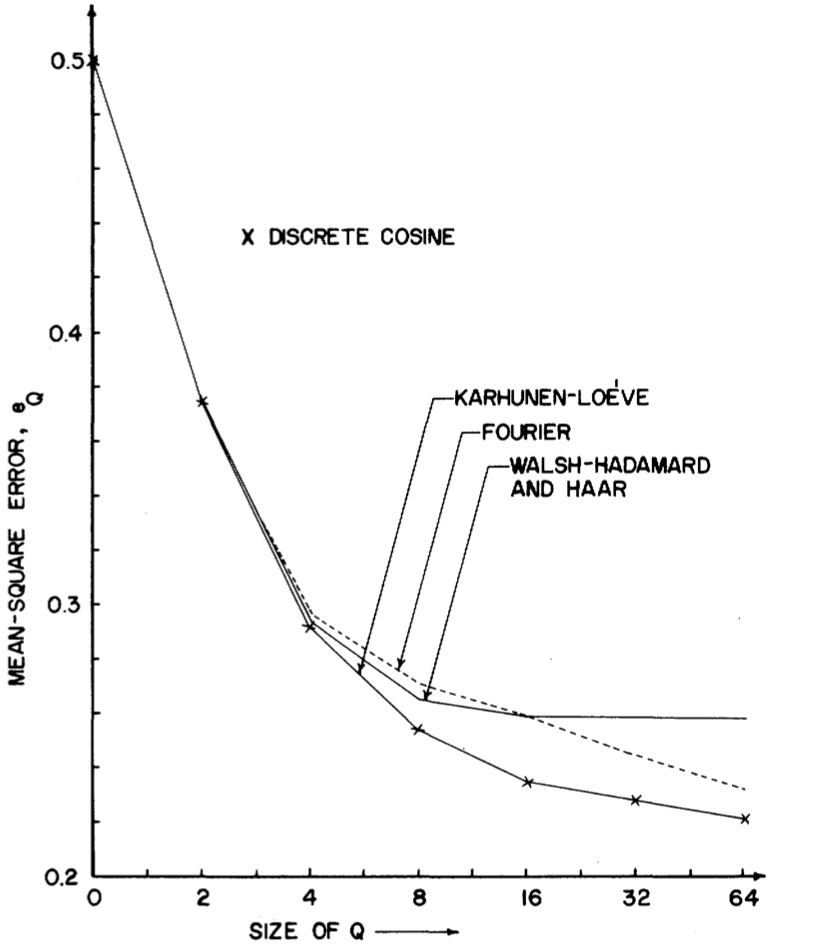}}
\caption{Scalar Weiner filtering, mean-Square error performance of Karhunen-Loève, Fourier, Walsh-Hadamard, and Haar transforms. Given $\rho=0.9$\cite{b7}}
\label{fig2}
\end{figure}

Figure \ref{fig2} also illustrates this close optimality between Karhunen-Loève transforms and DCT when comparing performance curves obtained from the mean-square error in scalar Weiner filtering with $\rho=0.9$.

We can also compute the DCT matrix for $N=4$. Using the formula given in \eqref{eq2} by expressing it in cosine form, equation \eqref{eq3} gives us the kernel for a 1D DCT matrix.

\begin{equation}
X[k]=\sum_{n=0}^{N-1}x[n]\cos\left[\frac{(2n+1)\pi k}{2N}\right]\label{eq3}
\end{equation}

where $N-1$. $0\leq k \leq N-1$

\begin{equation}
\alpha(k)=\begin{cases} 
\sqrt{\frac{1}{N}}\:\text{if}\:k = 0\\
\sqrt{\frac{2}{N}}\:\text{if}\:k \neq 0\\
\end{cases}\label{eq3}
\end{equation}

We thus obtain the following kernel after collecting the coefficients of $x(0),x(1),x(2)$ and $x(3)$ from $X[k]$ after substituting $N = 4$ into equation \eqref{eq3}. The transform is \textit{real} and \textit{orthogonal}.

\begin{equation}
\begin{bmatrix}
X[0]\\
X[1]\\
X[2]\\
X[3]\\
\end{bmatrix}=
\begin{bmatrix}
0.5 & 0.5 & 0.5 & 0.5\\
0.6532 & 0.2706 & -0.2706 & -0.6532\\
0.5 & -0.5 & -0.5 & 0.5\\
0.2706 & -0.6533 & 0.6533 & -0.2706\\
\end{bmatrix}
\begin{bmatrix}
x[0]\\
x[1]\\
x[2]\\
x[3]\\
\end{bmatrix} 
\label{eq4}
\end{equation}

Where, $M\times M^{T}=I $

Lee \cite{b8} proposes a novel approach to make the system more manageable by reducing the number of multiplications to about half of those required by the existing DCT algorithm initially proposed by Ahmed et al. Furthermore, DCT also has fast computation implementations available unlike KLT \cite{b12}. One such implementation is a fast-recursive algorithm proposed by Hseih in \cite{b11}. This method requires fewer multipliers and adders as it allows the ``generation of the next higher order DCT from two identical lower order DCTs." Other fast-recursive implementations are surveyed and considered in \cite{b13}.

In \cite{b14} and \cite{b15}, McMillan et al. propose a patent for fast implementation of discrete inverse cosine transform in digital image processing using low-cost accumulators or using optimized lookup tables.

\subsection{Joint Photographic Experts Group Compression}

JPEG compression \cite{b16} is a generic standard for the compression of grayscale and color, continuous-tone, still images. Typically preceded by the file extensions .jpg or .jpeg. This standard is categorized as image/jpeg under the MIME media type when uploading images. According to \cite{b17}, ``JPEG/JFIF supports a maximum image size of $65,535 \times 65,535$ pixels, hence up to 4 gigapixels for an aspect ratio of 1:1”.

Furthermore, a report published by \cite{b18} as of 2016 stated that JPEG is one of the most widely utilized common formats for storing and transmitting photographic images across the Internet. It aims to reduce transmission and storage costs. Furthermore, while also offering affordable image acquisition, display devices and ensuring interoperability among vendors. In fact, the ``joint” in JPEG refers to the collaboration between International Organization for Standardization (ISO) and Comité Consultatif International Téléphonique et Télégraphique  (CCITT)/International Telegraph Union-Telecommunication Standardization Sector (ITU-T) and hence JPEG is both an ISO Standard and CCITT recommendation.

The JPEG standard supports two compression methods – lossy compression using a DCT-based method and lossless compression using a predictive method. 

In lossless compression, there is almost no loss of information post compression. These techniques are always guaranteed to generate a replica of the original input image. The resultant file that is compression is a duplicate of the source file after the compress/expand cycle \cite{b19}. Such compression is typically used in mission-critical applications such as health and military database records, where even the mishandling of a single bit could be catastrophic. Run-Length encoding (RLE) and Lempel–Ziv–Welch (LZW) \cite{b20} compression techniques are some of the common methods for lossless compression. 

\begin{figure}[htbp]
\centerline{\includegraphics[width=0.5\textwidth]{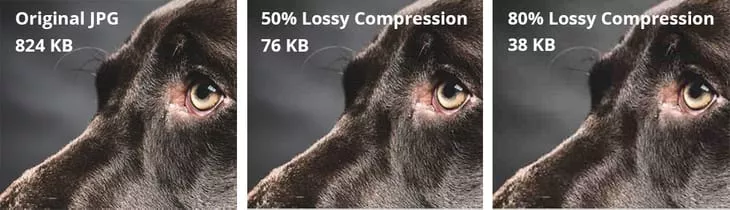}}
\caption{Image degradation with lossy compression at higher compression ratios \cite{b21}}
\label{fig3}
\end{figure}

In lossy compression, on the other hand, the decompressed image is not an exact match of the source input image. Information loss is apparent but aids in the reduction of file sizes. Furthermore, the lost information cannot be restored and results in loss of image quality. Figure \ref{fig3} depicts this slight loss of image quality of an image at different compression ratios. Hence, this form of compression is also termed as irreversible compression. The compression ratios are higher in lossy compression methods due to the loss of information, e.g., it can be as low as 50:1. Thus, such methods are used in scenarios such as archiving data where an exact reproduction of the original data is not vital. Examples of lossy compression methods are JPEG, JPEG 2000 \cite{b22} and Wavelet Compression \cite{b23}. 

The following are the goals as defined by JPEG before synthesizing an architecture for JPEG image compression:

\begin{itemize}
\item Provide “very good” or “excellent” image quality rating post compression.
\item The encoder should provide flexibility to the user or application to select the desired compression and/or quality tradeoff. 
\item Should not be restricted or biased to any towards any image attributes, this includes, but is not limited to, dimensions, aspect ratio, range of colors and statistical properties.
\item The JPEG algorithm must have a tractable computational complexity or be solvable in polynomial time. Thus, making it feasible to create software implementations that are cost effective and run on a wide range of single processor CPUs.
\end{itemize}

JPEG also aims to provide the following four modes of operation:

\begin{enumerate}
\item Sequential encoding – images are encoded in a single top to bottom or right to left scan.
\item Progressive encoding – encoding takes place in multiple scans for applications with prolonged transmission time.
\item Hierarchical encoding – the image is encoded at multiple resolutions. This way several resolutions, even lower ones, are obtainable without having to decompress the source image.
\item Lossless encoding – the image is encoded in such a manner that guarantees recovery of an exact copy of the original image with no information loss.
\end{enumerate}

\section{DCT In JPEG Compression}

The basis for all DCT-based decoders is the Baseline sequential codec. Equation \eqref{eq1} previously illustrated the formula for Forward DCT (FDCT) or DCT image compression. Equation \eqref{eq5}, derived from \eqref{eq1}, illustrates the mathematical definition for Inverse DCT (IDCT) used in a DCT-based decoder.

\begin{multline}
f[m,n]=\sum_{k=0}^{N-1} \sum_{l=0}^{N-1}\alpha(k)\alpha(l)F(k,l) \\
\cos\left[\frac{(2m+1)\pi k}{2N}\right]\cos\left[\frac{(2n+1)\pi l}{2N}\right] \label{eq5}
\end{multline}
where,
\begin{equation}
\alpha(k)\: \text{and}\: \alpha(l)=\begin{cases} 
\sqrt{\frac{1}{N}}\:\text{if}\:k = 0\\
\sqrt{\frac{2}{N}}\:\text{if}\:k \neq 0\\
\end{cases}\label{eq5}
\end{equation}

Given the ``four modes of operation” in the previous section, a codec is specified for each version. The reader must note that though codec is being used in several times in this assignment, it is not mandatory for applications to implement both an encoder and decoder. JPEG and DCT’s flexibility allows for the independent implementation of each operation mode’s decoder and encoder separately. 

\begin{figure}[htbp]
\centerline{\includegraphics[width=0.5\textwidth]{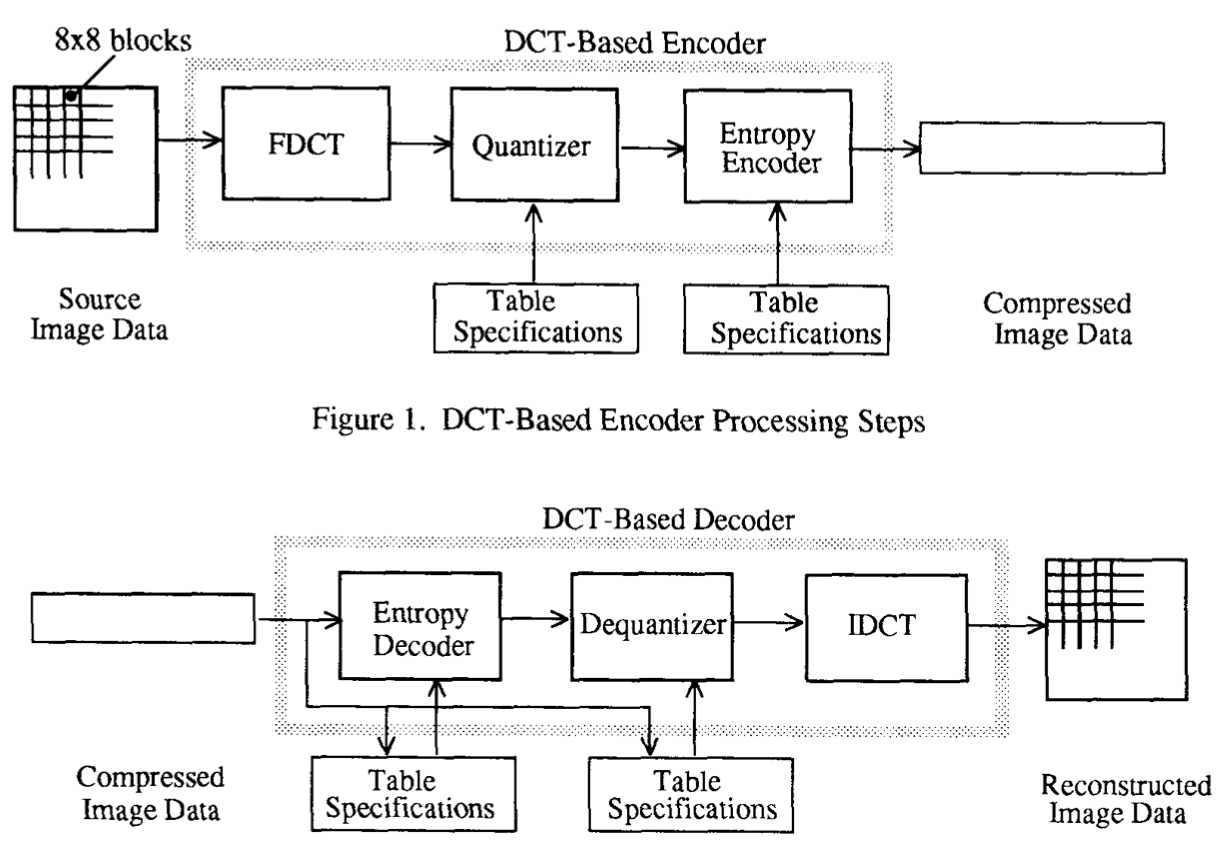}}
\caption{(a, top) processing steps for a DCT-based Encoder and (b, bottom) a DCT-based Decoder  \cite{b16}}
\label{fig4}
\end{figure}

For a baseline sequential codec operation mode, figure \ref{fig4} (a) and (b) illustrate the compression in a complete manner. While for a progressive-mode codec, prior to the entropy coding step, an image buffer must be maintained. This allows for multiple scans with successive improvements between consecutive scans. Furthermore, hierarchical-mode codec uses a broader and more comprehensive framework while borrowing aspects from the process described in figure \ref{fig4}.

Assuming an interleaving of 8 × 8 blocks (single-component, grayscale images) as inputs for the FDCT, 64 DCT coefficients are produced. These coefficients are then used in the quantization process as shown in figure \ref{fig4} (a). 

\begin{figure}[htbp]
\centerline{\includegraphics[width=0.5\textwidth]{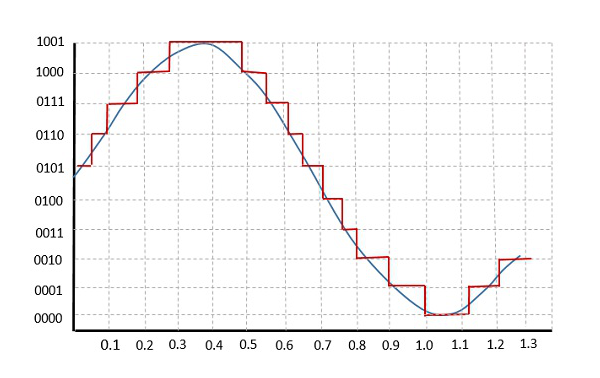}}
\caption{A simple quantization of a signal by choosing the amplitude values closest original analog amplitude  \cite{b25}}
\label{fig5}
\end{figure}

Quantization, as described in \cite{b24}, is the processes of mapping a set of larger input values to a smaller set of output values, almost analogous to the process of rounding off numbers. The above-described implementation of a DCT-based encoder utilizes a 64-element quantization table taken as an input from the user. This process involves digitizing the amplitude or brightness values. The red line in figure \ref{fig5} depicts this digitizing an analog signal. The discrete amplitude of the quantized output is represented on the y-axis as a binary digit. These discretized levels are known as representation levels separated by spacing called quantum. The signal is approximated to the closest level. This process is the core of all lossy compression algorithms and helps in achieving further compression. 

In the given implementation, quantization is performed in conjunction with a quantization table and the input signal is digitized. This process is fundamentally lossy since this is a many-to-one mapping. This method causes the image loss in lossy image compression for DCT-based encoders. Quantization can be represented using equation \eqref{eq6}. This is done by rounding of the quotient of dividing each DCT coefficient by its corresponding quantum to the nearest integer. Followed by normalization of $F_Q (k,l)$.

\begin{equation}
F_Q (k, l)= \text{Integer Round}\left(\frac{F(k,l)}{Q(k,l)}\right)
\label{eq6}
\end{equation}

Normalization is removed via multiplication with the quantum in IDCT, thus, returning the original coefficient supplied for quantization during DCT. This is also represented as equation \eqref{eq7}.

\begin{equation}
F_{Q'} (k, l)= F_Q (k,l)\times Q(k,l)
\label{eq7}
\end{equation}

Some quantization tables can be referred from \cite{b26} for CCIR-601 images and displays. Furthermore, they are ISO and JPEG recommendations, but not requirements.

The final step is entropy coding in DCT and its counterpart, entropy decoding corresponds to the first step in IDCT. The DC coefficients are taken separately from the AC coefficients after quantization as given in figure 6. The difference of the DC terms in the previous block in the encoding order as shown in figure 6. Followed by arranging the DC coefficients in a zig-zag sequence to help facilitate entropy coding.

\begin{figure}[htbp]
\centerline{\includegraphics[width=0.5\textwidth]{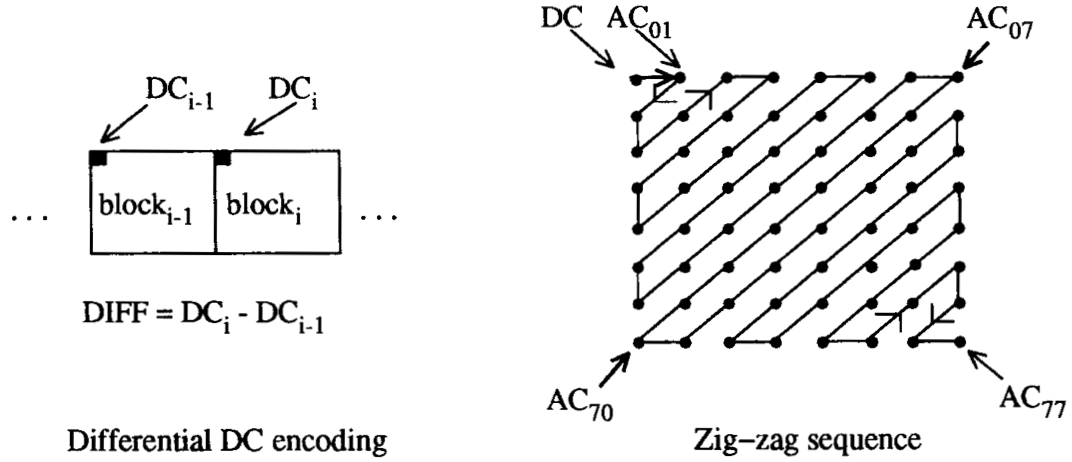}}
\caption{DC coefficients and AC coefficients being treated separately for entropy coding.  \cite{b16}}
\label{fig6}
\end{figure}

JPEG recommends two encoding methods for entropy coding – Huffman coding \cite{b27}, primarily used by baseline sequential codec, and arithmetic coding \cite{b28}. However, both codecs can be used for all modes of operation. Using the statistical characteristics of the coefficients, these entropy coding methods attain additional compression losslessly. This is in contrast to quantization, where some of the data is lost.

Entropy coding is defined as a two-step process by JPEG:
\begin{enumerate}
\item The zig-zag sequence (figure \ref{fig6}) is converted into an intermediate sequence of symbols.
\item The symbols are then converted into a data stream with no `externally identifiable boundaries.'
\end{enumerate}
Each of the coding methods specify the form and definition of the intermediate symbol.
Though arithmetic coding can be seen as a more complex process than Huffman coding, it produces 5-10

\section{Other Applications of DCT}

As discussed in the previous section, DCT finds its primary application in lossy image compression. This is due to its strong “energy compaction” property and its closely related performance to KLT for strongly correlated Markov processes \cite{b7}\cite{b29}. 

A variant of DCT, Modified Discrete Cosine Transform (MDCT) \cite{b30}, and has found implications for audio coding and error concealment \cite{b31}. MDCT typically are designed using a recursive DCT-IV algorithm. Furthermore, audio formats such as MP3 \cite{b32}, AAC \cite{b33}, WMA and Vorbis \cite{b34} employ MDCTs for audio compression.  This is because the data throughput for MDCT and Inverse MDCT algorithms is four times higher than previous algorithms. Furthermore, it also boasts a 50\%-79\% in ROM size. Hence, increasing overall chip efficiency and providing feasible architectural solutions.

Multidimensional DCTs or MD DCTs, another variant of DCT, finds its application in video compressions such as Theora video compression \cite{b35}, MPEG and Daala. More advanced applications include adaptive video encoding \cite{b36}; this implementation uses an MD DCT coder for medical images. The MD DCT or 3D DCT is used to compress 3D cuboid which is a resultant of the given segmentation technique.

\section*{Acknowledgment}

The author, Jacob John would like to thank Dr. Prabu Sevugan for his continuous support throughout this paper. I would also like to thank Vellore Institute of Technology for their aid without which this paper wouldn't have been completed.

\end{document}